\documentclass[conference]{IEEEtran}
\IEEEoverridecommandlockouts

\usepackage{cite}
\usepackage{amsmath,amssymb,amsfonts}
\usepackage{algorithmic}
\usepackage{graphicx}
\usepackage{textcomp}
\usepackage{xcolor}

\usepackage{hyperref}
\usepackage{balance}
\usepackage{fancyhdr}
\usepackage{tabularx}
\usepackage{lipsum}
\usepackage{graphicx}
\usepackage{subfigure}
\usepackage{enumitem}
\usepackage{wrapfig}
\usepackage{amssymb}
\usepackage{pifont}
\usepackage[ruled,vlined]{algorithm2e}
\usepackage{mathtools}

\usepackage{amsmath}
\usepackage{soul}
\usepackage{multirow}
\usepackage[square, numbers]{natbib}
\usepackage{wrapfig}
\usepackage{booktabs}
\usepackage{colortbl}

\newcommand{\cmark}{\ding{51}}%
\newcommand{\xmark}{\ding{55}}%
\newcommand{\ds}[1]{LAVN}

\def\BibTeX{{\rm B\kern-.05em{\sc i\kern-.025em b}\kern-.08em
    T\kern-.1667em\lower.7ex\hbox{E}\kern-.125emX}}

\begin{document}

\title{A Landmark-Aware Visual Navigation Dataset for Map Representation Learning
}

\author{\IEEEauthorblockN{Faith Johnson\textsuperscript{*}\thanks{\textsuperscript{*}Both authors contributed equally to this work.}, Kristin Dana}
\IEEEauthorblockA{\textit{Rutgers University} \\
New Brunswick, NJ, USA \\
\{faith.johnson,kristin.dana\}@rutgers.edu}
\and
\IEEEauthorblockN{Bryan Bo Cao\textsuperscript{*}, Shubham Jain}
\IEEEauthorblockA{\textit{Stony Brook University} \\
Stony Brook, NY, USA \\
\{boccao,jain\}@cs.stonybrook.edu}
\and
\IEEEauthorblockN{Ashwin Ashok}
\IEEEauthorblockA{\textit{Georgia State University} \\
Atlanta, GA, USA \\
{aashok@gsu.edu}}
}




\maketitle

\begin{abstract}
\label{sec:abs}
Map representations learned by expert demonstrations have shown promising research value. However, the field of visual navigation still faces challenges due to the lack of real-world human-navigation datasets that can support efficient, supervised, representation learning of environments. We present a \textbf{L}andmark-\textbf{A}ware \textbf{V}isual \textbf{N}avigation (LAVN) dataset to allow for supervised learning of human-centric exploration policies and map building. We collect RGBD observation and human point-click pairs as a human annotator explores virtual and real-world environments with the goal of full coverage exploration of the space. The human annotators also provide distinct landmark examples along each trajectory, which we intuit will simplify the task of map or graph building and localization. These human point-clicks serve as direct supervision for waypoint prediction when learning to explore in environments. Our dataset covers a wide spectrum of scenes, including rooms in indoor environments, as well as walkways outdoors. We release our dataset with detailed documentation at \href{https://huggingface.co/datasets/visnavdataset/lavn}{\textcolor{blue!65}{https://huggingface.co/datasets/visnavdataset/lavn}} (\textcolor{blue!65}{DOI: 10.57967/hf/2386}) and a plan for long-term preservation.
\end{abstract}


\section{Introduction}
Robotic navigation remains a research problem of interest as advances in AI make intelligent mobile agents, such as delivery bots, more feasible. Many approaches have been proposed from SLAM \cite{mur2015orb, mur2017orb, campos2021orb}, to graph-based methods \cite{bavle2023faster, bavle2022s, kim2023topological, he2023metric}, to reinforcement learning (RL) \cite{zhu2017target} to enable robots to effectively move around a space in a goal-oriented 
manner. However, these methods suffer from high computational complexity, high memory usage, and low sample efficiency respectively. Visual navigation \cite{ramakrishnan2022poni,morin2023one,hahn2021nrns} methods circumvent several of 
these challenges by using neural networks to implicitly learn scene geometry and an optimal navigation policy without necessitating direct environment interaction.

\begin{figure}[t]
  \centering
  \includegraphics[width=0.75\linewidth]{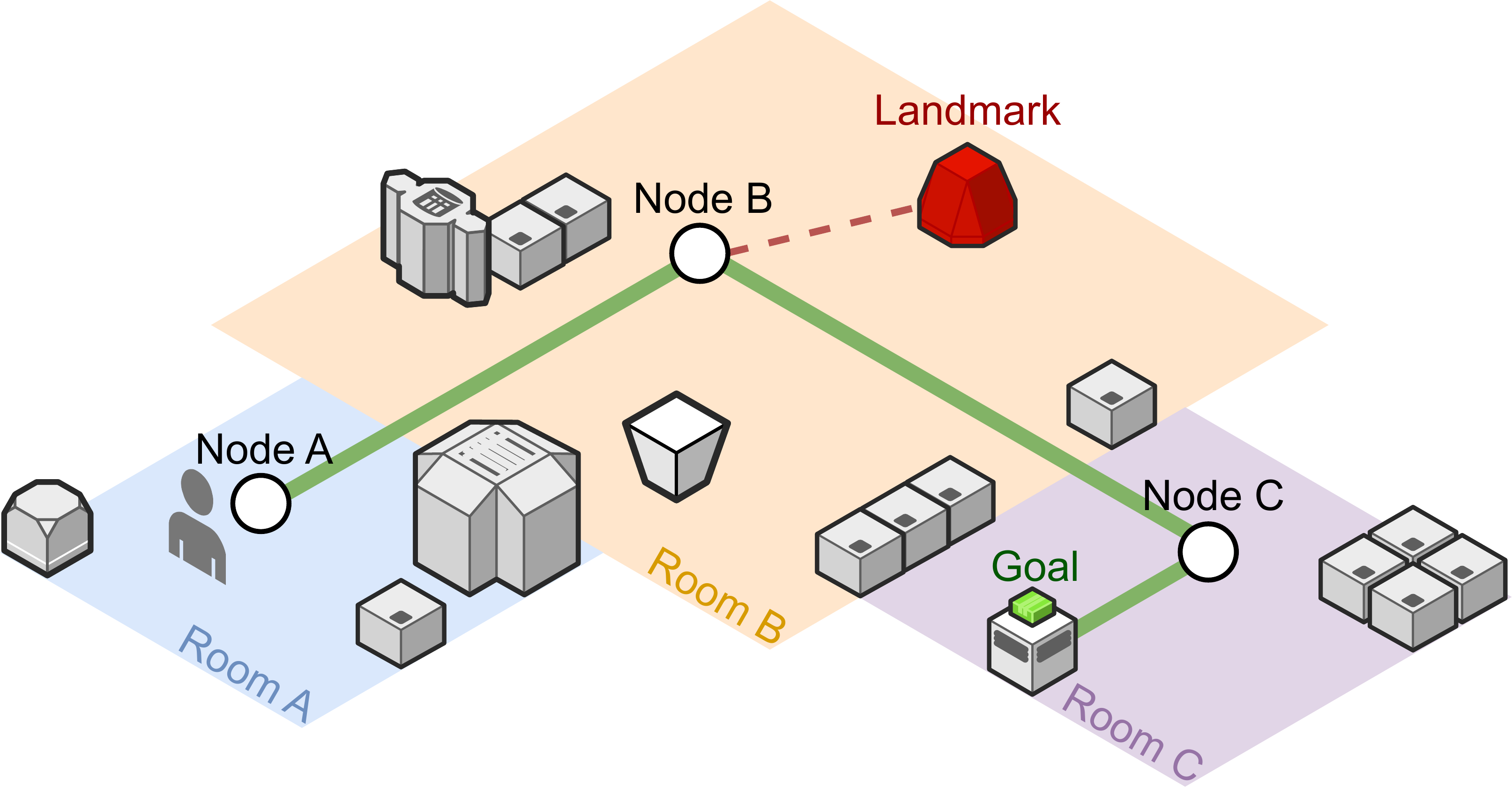}
  \caption{We propose a novel Landmark-Aware Visual Navigation (LAVN) dataset for map representation research. LAVN provides unique landmark information (in red) in specific scenes to aid model navigation and exploration performance and map representation learning. In this figure, a human expert in Room A (in blue) provides a video-based training trajectory (in green lines) for navigation given a goal image (green box). Rooms are depicted in different colors and represented with nodes. Our dataset facilitates topological map representation learning by injecting landmarks into a robot's observation.}
  \label{fig:dataset}
\end{figure}

Typically, visual navigation methods are evaluated with the image goal task, where an agent is given a goal image of some target and tasked with finding and reaching it in the environment. 
The success of state of the art (SOTA) methods at this task comes down to an agent's ability to efficiently explore its environment. Several works focus on explicitly learning optimal exploration policies \cite{chaplot2020learning,ye2021auxiliary,chaplot2021seal,chaplot2020object}. However, they either abstract exploration out of the problem all together by assuming that the environment has been seen by the agent prior to test time \cite{morin2023one}, with the hope that their networks will implicitly learn an adequate exploration strategy from the data due to the specific task structure \cite{krantz2021waypoint}, or explicitly define the exploration strategy for the agent by hard coding it into the architecture \cite{hahn2021nrns}.

Of those methods that learn to explore, several niches have arisen such as adding an exploration term to RL reward functions with \cite{ye2021auxiliary} and without \cite{chen2019learning} behavioral cloning, building metric maps or graphs with \cite{chaplot2021seal} and without \cite{chaplot2020neural} semantics, and learning to propose exploration waypoints that provide helpful subgoals during navigation for local policies \cite{chaplot2020object}. These methods suffer from similar downsides as those mentioned previously with the added potential for also requiring full ground truth semantic labels of an environment, which can be extremely costly to acquire.   
Instead, a much simpler approach would be to explicitly learn good exploration techniques in a human-centric manner. 
Learning human-centric exploration strategies in a supervised manner would allow for much simpler networks to effectively solve navigation problems, thus reducing the computational complexity, high memory usage, and low sample efficiency issues of SOTA. However, there is very limited data that would allow for this kind of learning. 

In this work, we present a \textbf{L}andmark-\textbf{A}ware \textbf{V}isual \textbf{N}avigation (LAVN) dataset of human-guided environment exploration videos collected from the photorealistic Habitat virtual environment as well as several real world environments. We collect RGB image observations and human point click pairs as a human guides an agent in exploring each environment. These point clicks represent subgoals that guide the agent incrementally though the space and can be used for supervised learning of human-centric waypoints for environment exploration. Additionally, we expect that having landmarks when exploring has the potential to improve current SOTA performance on the map-building and localization tasks inherent in navigation. To this end, the human annotator also provides examples for helpful navigational landmarks, which will allow for supervised training of a landmark prediction network as well. In this way, it becomes possible to directly train a network to explore an environment in a human-centric manner under multiple SOTA paradigms as demonstrated by the recent work \cite{johnson2024feudal, johnson2024memory} in the visual navigation task. In addition, learning from such waypoint and landmark annotations allows for aligning the cognitive mindset of human navigation with mobile robots, enabling intuitive interactions with people to convey mobile robots' intent in navigation {via gaze behavior generation} \cite{ranganeni2023exploring, yu2023your, bevins2024user, fujioka2024need, huang2024gestures}.

This paper is organized as follows: we summarize our dataset's key differences from existing works in Section \ref{sec:relatedwork}. Then we present the details of data acquisition both in the virtual and real-world environments in Section \ref{sec:vd} and \ref{sec:rd}, respectively. We conclude our work in Section \ref{sec:concl}.


\section{Related Work}
\label{sec:relatedwork}


\subsection{Map Representation for Navigation}
The success of navigation relies on an effective learning and understanding of the environment. Two main approaches for this are map representation learning in geometry-based \cite{mur2015orb, mur2017orb, campos2021orb} and learning-based \cite{chaplot2020learning, zhu2017target, morin2023one, shah2021ving, vezhnevets2017feudal} domains. However, the former requires careful and robust hand-engineered feature design and requires intensive computation (e.g. keyframes in SLAM). The advancements of Deep RL in the second category are partially attributed to the emergence of simulators designed for photorealistic simulation, including Habitat AI \cite{szot2021habitat, habitat19iccv} with datasets Gibson \cite{xiazamirhe2018gibsonenv} and Matterport \cite{Matterport3D}. While these simulators enable large scale model training, they are also expensive to build, can run slowly enough to mitigate their ability to produce large numbers of training samples, and introduce the sim2real problem when deploying models in the real world. To combat this, methods like NRNS \cite{hahn2021nrns} demonstrate the effectiveness of using expert videos to train robots without interacting with environments.

\subsection{Human-Robot Interaction Datasets}
There is a large body of datasets or benchmarks for Human-Robot Interaction, ranging from industrial \cite{iodice2022hri30} to social \cite{karnan2022socially, rudenko2020thor, day2023study, wan2022handmethat, anvari2020modelling, schreiter2022magni, shu2016learning, chen2022human} environments. As one of the fundamental tasks for robots, researchers have collected datasets for learning visual navigation with socially compliant behaviors \cite{karnan2022socially, rudenko2020thor, day2023study, anvari2020modelling, schreiter2022magni, shu2016learning, chen2022human, ranganeni2023exploring, bevins2024user}, including intuitively conveying mobile robot's navigation intent \cite{yu2023your, fujioka2024need} to humans. Apart from learning from other robots collectively \cite{shah2023gnm}, robotic visual navigation can benefit from human feedback via visual cues \cite{wignessrugd}, language \cite{thomason2020vision}, gesture \cite{cao2018diffframenet, huang2024gestures}, multimodal instructions \cite{schreiter2022magni, shrestha2023natsgd}, or demonstrations \cite{ramrakhya2022habitat}.

Our work extends the line of efficient map representation using topological graphs \cite{he2023metric, kim2023topological}. Similar to how a human \cite{chrastil2014cognitive} can quickly learn to navigate in a new environment by implicitly treating landmarks as nodes in their mental graph \cite{chrastil2014cognitive}, we hypothesize that a robot can acquire the ability to navigate in novel places efficiently by intelligently injecting landmarks into its own topological graph. However, little attention has been paid to such datasets, resulting in \textbf{a lack of data} that includes \textbf{landmark} information for this research in the literature. Our dataset hereby aims to fill this gap by providing a series of human-expert-annotated videos with \textbf{landmarks} both in virtual and \textbf{real-world} environments, including indoor and outdoor scenes. We summarize the key characteristics of recent datasets in Table \ref{tab:dsrelated}.

\begin{table}[t]
  \begin{center}
\caption{Summary of dataset characteristics for visual navigation. We provide a unique dataset with beneficial navigational landmark labels in human trajectory demonstrations in both real world and simulated environments. Add.: Additional; Mod: Modality; Sem: Semantics; LM: Landmark.}
\begin{tabular}{c|ccccccc}
\toprule
\multirow{3}{2.6em}{Dataset} & \multirow{3}{2em}{\centering Human\\ Traj.\\ Demo} & \multirow{3}{1.2em}{\centering Sim} & \multirow{3}{1.2em}{\centering Real} & \multirow{3}{1.2em}{\centering Small\\ Scale} & \multirow{3}{1.2em}{\centering No\\ Add. \\ Mod.} & \multirow{3}{1.2em}{\centering No \\ Sem.} & \multirow{3}{1.2em}{\centering LM}\\
&  & & & &  & \\
&  & & & &  & \\
\midrule
Habitat-Web \cite{ramrakhya2022habitat} & \cmark & \cmark & \xmark & \xmark & \cmark & \cmark & \xmark \\
RUGD \cite{wignessrugd} & \cmark & \xmark & \cmark & \cmark & \cmark & \xmark & \xmark \\
RealEstate20k \cite{zhou2018stereo} & \cmark & \xmark & \cmark & \cmark & \cmark & \cmark & \xmark \\
EmbodiedQA \cite{das2018embodied} & \cmark & \cmark & \xmark & \cmark & \xmark & \cmark & \xmark \\
SCAND \cite{karnan2022socially} & \cmark & \xmark & \cmark & \cmark & \cmark & \cmark & \xmark \\
\rowcolor{gray!30} \textbf{LAVN (Ours)} & \cmark & \cmark & \cmark & \cmark & \cmark & \cmark & \cmark \\
\bottomrule
\end{tabular}

\end{center}
\label{tab:dsrelated}
\vspace{-5pt}
\end{table}





\begin{figure*}[h]
  \centering
    {\includegraphics[width=0.93\textwidth]{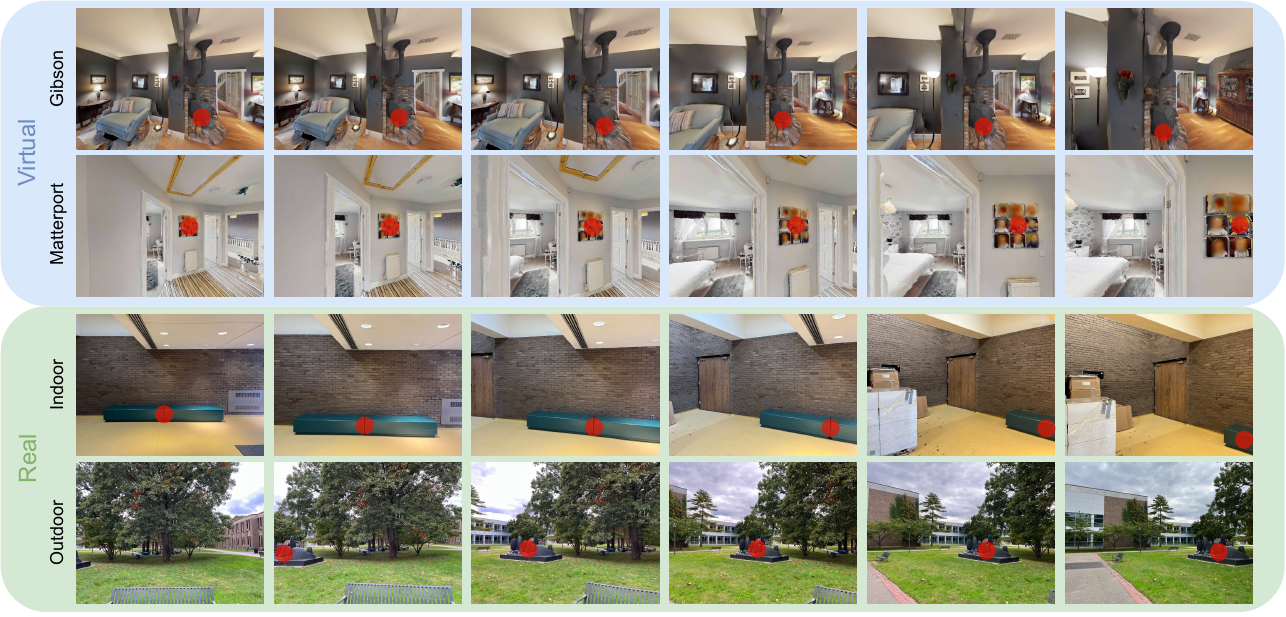}}
  \caption{Samples of RGB images in our dataset captured from an ego-centric camera. Video frames start from left to right. Landmarks are visualized with red dots. Samples in the virtual dataset (1st and 2nd rows in blue) are visually realistic compared to the real-world observations (3rd and 4th rows in green), which makes landmark data acquisition scalable to serve \textbf{real-world} research purposes.}
  \label{fig:vis}
  \vspace{-5pt}
\end{figure*}


\section{Virtual Dataset}
\label{sec:vd}
The main objective of the data collection is to provide human navigation demonstrations for robots in a real-world environment. As video has been proven to be an effective way to teach robots to accomplish different tasks \cite{hahn2021nrns}, we hereby collect trajectories in video format. Below, we detail the process of data collection for the virtual dataset.



\subsection{Environments}
We provide human-click annotations for a selection of the Matterport \cite{Matterport3D} and Gibson environments \cite{xiazamirhe2018gibsonenv} in Habitat-Sim \cite{szot2021habitat,habitat19iccv} to facilitate virtual environment exploration. We provide a full list of annotated environments with the data. Habitat is already used to train many visual navigation methods, so having in-domain, landmark-aware data will allow for direct extensions of these methods for greater ease of model comparison. Additionally, it allows for greater data collection volume and lower data processing time.

{We aim at randomly selecting rooms evenly distributed by room IDs to ensure diversity in room types and sizes, as detailed in the Appendix.}


\subsection{Data Collection and Representation}
\label{ds:vir}
To start, the human navigator is placed at a random, navigable point in an environment, and tasked with visiting as many rooms as possible while identifying distinct landmarks in each room. The current first person observation of the environment is shown on a screen, and the human moves through the environment by left clicking on this image towards the direction they would like to move. There are three discrete moving action choices: moving forward 0.25m, turning left $15^{\circ}$, and turning right $15^{\circ}$. To specify a landmark, the navigator right clicks on the observation image over the object or region of space they choose. We provide a human point-click supervisory signal for each time step of the trajectory, but these signals could easily be masked to generate data to train temporally abstracted waypoint predictors. 

A graph is built as the navigator moves through the space, where nodes are added each time an action is taken. Each node contains one RGB observation, one depth image, the human navigational point click, and the ground truth odometry (including ground truth location and orientation). Edges are drawn between consecutive nodes only, however it is possible to perform loop closure on the graph using the odometry provided. Each graph also contains a sequential list of the actions taken to move through the environment. {A landmark is defined as a unique and distinct object, distinguishable from other common objects in the scene.} A list of pairs of  landmark point-click image coordinates and their corresponding nodes is also provided. Each human-guided trajectory is terminated after 500 actions or when the annotator has navigated through the entirety of the space, whichever comes first. 

\begin{figure*}[h]
  \centering
  \includegraphics[width=0.93\linewidth]{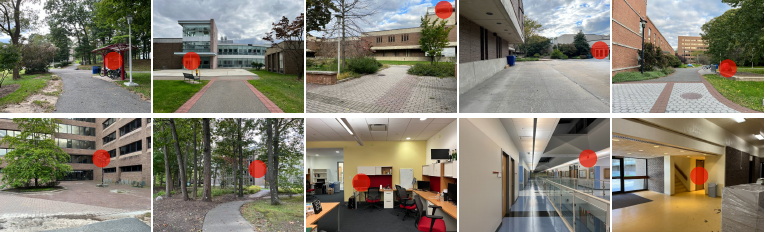}
  \caption{Samples of RGB images in the real world environments with landmarks annotated by red dots. Trajectories are recorded in comprehensive environments both indoors and outdoors.}
  \label{fig:dataset}
  \vspace{-5pt}
\end{figure*}

\section{Real World Dataset}
\label{sec:rd}

\subsection{Environments}
Real-world data is recorded in a broad spectrum of environments. Indoor scenes include laboratories and hallways inside a building, while outdoor environments contain avenues, walkways, and campus buildings with different architectures. A wide range of trajectories with various complexities are captured in both regular shapes such as rectangles in corridors inside a building, as well as irregular shapes on roads surrounded by trees or near a lake. Objects in common environments are included, ranging from chairs, desks, doors, and computers in indoor rooms, to trees, lakes, stones, bikes, bricks and so forth in open spaces outside a building.

\subsection{Data Collection and Representation}
In order to collect consistent data between the real-world data and simulation, the human annotator is asked to follow the same protocol as in the virtual environment data collection described in Section \ref{ds:vir}. During navigation, the annotator holds a camera (iPhone13Mini \cite{iphone13mini}'s built-in camera) naturally around 1.5m high off the ground facing forwards. Unlike in simulation where agents can move with exact distances or turning angles, real-world environments require that each action taken by the human navigator is inherently inexact. 
Therefore, the navigator is instructed to follow the same discrete action choices by their own judgement. Specifically, the expert takes a picture (saved in JPEG) in every step (around 0.7m), or in every left or right turn (roughly $15^{\circ}$). This also presents the real-world data with complex noise or perturbations, which can improve downstream method robustness.

\begin{figure}[h]
  \begin{center}
  \centering
  \subfigure[Frames]{\includegraphics[width=0.45\linewidth]{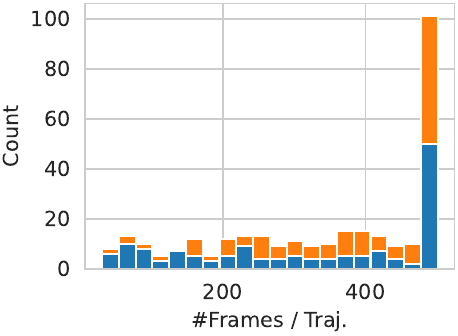}}\label{fig:f}
  \hspace{12pt}
  \subfigure[LM]{\includegraphics[width=0.45\linewidth]{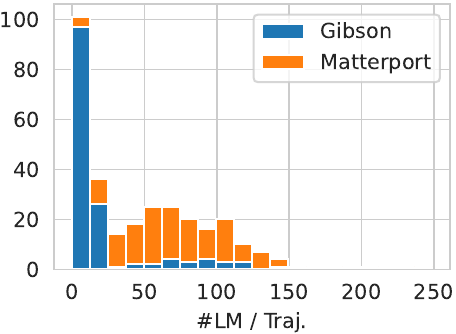}}\label{fig:lm}
  \caption{Stacked histograms depicting $\#$Frames and $\#$LM (landmarks) per trajectory. (a) The majority of trajectories consists of the maximum $\#$Frames 500. (b) Most trajectories in the Gibson rooms consists of only a small number of annotated 
  landmarks (e.g. $<25$), while it increases in Matterport due to its larger and more complicated environments. 
  }
  \label{fig:datahist}
  \end{center}
  \vspace{-5pt}
\end{figure}


\begin{table}[h]
  \begin{center}
  \caption{Statistics of the LAVN dataset.}
    {\small{
\begin{tabular}{ccccccc}
\toprule
$\#$Traj. & Total & Total & $\#$Virt. & $\#$Real World\\
& $\#$Frames & $\#$Landmarks & Envs & Envs\\
\midrule
310 & 103,998 & 14,281 & 300 & 10\\
\bottomrule
\end{tabular}
}}
\end{center}
\label{tab:stats}
\vspace{-5pt}
\end{table}

Resolution in the original images was 4032x3024. To promote consistency between the virtual and real-world data, the images are resized to be the same resolution of 640x480 as in the simulation and saved in JPEG format with quality $85\%$. Landmarks are annotated offline after the videos are saved. Samples from our dataset are visualized in Fig. \ref{fig:vis}. Both virtual and real-world images are photorealistic and can benefit downstream research in the real world. Also notice that trajectories are recorded in ten diverse scenes in both indoor and outdoor environments shown in Fig. \ref{fig:dataset}. Our proposed dataset provides a unique benchmark for both learning and evaluation to serve the visual navigation community. Our dataset statistics are summarized in Table \ref{tab:stats} and Fig. \ref{fig:datahist}. Observe that most trajectories reach the maximum number of frames (500). On the other hand, a wide variety of  trajectory horizon lengths ($\#Frames / Traj.$) and number of landmarks ($\#LM / Traj.$) is covered which benefit robot learning in both short \cite{wang2020long} and long horizon \cite{meng2021learning, pertsch2020long, fang2019scene} tasks.

\begin{figure}[h]
  \begin{center}
  \centering
  \subfigure[Action Stats]{\includegraphics[width=0.49\linewidth]{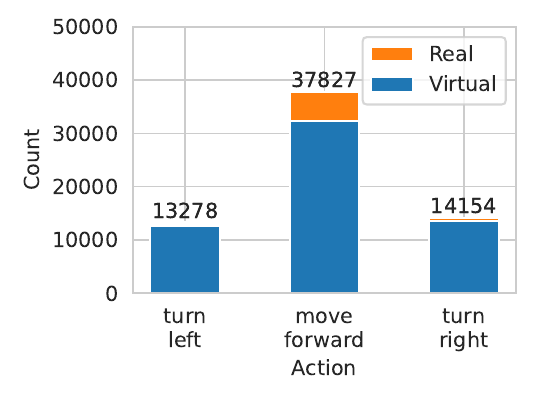}}\label{fig:act}
  \subfigure[Waypoint Density Map]{\includegraphics[width=0.49\linewidth]{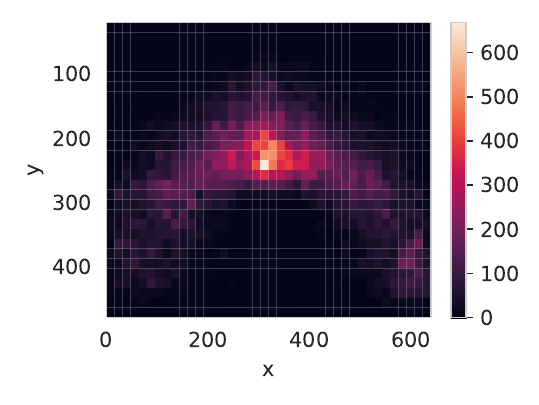}}\label{fig:wp}
  \caption{(a) Stacked bar chart of actions in virtual and real-world environments. (b) Waypoints heatmap over the entire dataset.
  }
  \label{fig:act_waypoints_stats}
  \end{center}
  \vspace{-5pt}
\end{figure}

{For both datasets, the graph is saved as a json file. We detail the file organization and examples of dataset usage in the Appendix.}

{\textbf{Real-world Relevance.} The photorealistic images from virtual datasets (Gibson and Matterport), combined with our real dataset, support robot visual navigation learning 
tailored for real-world deployment.}

\section{Conclusion}
\label{sec:concl}
In this work, we introduce the first \textbf{L}andmark-\textbf{A}ware \textbf{V}isual \textbf{N}avigation (LAVN) dataset with human point-click annotations, providing exploration waypoint and environment landmark labels that allow for supervised training for exploration and map building tasks. We provide trajectories in 300 virtual and real-world environments with RGB observations, depth images, odometry, and human point-clicks for navigation waypoint supervision and environment landmark supervision. With this dataset, it will be possible to {improve} visual navigation models that exhibit human-centric exploration policies on the image-goal task, paving the way for autonomous embodied agents in real-world applications. {Future work includes incorporating dynamic objects in the scene.}

\section*{Acknowledgment}
{This research was supported by the National Science Foundation (NSF) NRT NRT-FW-HTF: Socially Cognizant Robotics for a Technology Enhanced Socienty (SOCRATES), No. DGE-2021628 and Grant Nos. CNS-2055520, CNS-1901355, CNS-1901133.}





\begingroup
\footnotesize
\bibliographystyle{IEEEtran}
\balance
\bibliography{main}
\endgroup

\newpage
\onecolumn
\section{Dataset documentation and intended uses}
The datasheet for our dataset can be found at \href{https://huggingface.co/datasets/visnavdataset/lavn}{https://huggingface.co/datasets/visnavdataset/lavn} along with the data. We include samples from each trajectory in the supplemental material, available in the LAVN\_Samples.zip file.

{The intended uses of this dataset include, but are not limited to}:
\begin{itemize}
    \item image representation learning
    \item visual landmark prediction
    \item human navigation policy learning
    \item offline reinforcement learning
    \item robot navigation intent visualization
\end{itemize}

{The intended practical applications of this dataset include, but are not limited to}:
\begin{itemize}
    \item last-mile delivery
    \item factory/warehouse/hospital item transportation
    \item robotic assistance in retail environments
    \item search-and-rescue operations in hazardous terrains
\end{itemize}

\section{Accessing the Data}
The dataset is available at \href{https://huggingface.co/datasets/visnavdataset/lavn}{https://huggingface.co/datasets/visnavdataset/lavn}. Zipped files can be downloaded at \\ \href{https://huggingface.co/datasets/visnavdataset/lavn/tree/main}{https://huggingface.co/datasets/visnavdataset/lavn/tree/main}.

\section{Croissant Metadata}
The croissant metadata can be found at \href{https://huggingface.co/api/datasets/visnavdataset/lavn/croissant}{https://huggingface.co/api/datasets/visnavdataset/lavn/croissant}.

\section{Author Statement of Responsibility}
We, the authors, take full and total responsibility in case of violation of rights inherent in our dataset submission.

\section{Hosting, licensing, and maintenance plan.}
We choose to host our dataset on huggingface, where we detail the maintenance plan, licensing details (MIT license in this case), and other relevant information. We will conduct a long-term maintenance plan to ensure the accessibility and quality for future research:

\begin{itemize}
    \item \textbf{Data Standards}: Data formats will be checked regularly with scripts to validate data consistency.
    \item \textbf{Data Cleaning}: Data in incorrect formats, missing data or contains invalid values will be removed.
    \item \textbf{Scheduled Updates}: We set up montly schedule for data updates.
    \item \textbf{Storage Solutions}: HuggingFace, with DOI (doi:10.57967/hf/2386), is provided as a public repository for online storage. A second copy will be stored in a private cloud server while a third copy will be stored in a local drive.
    \item \textbf{Data Backup}: Once one of the copies in the aforementioned storage approach is detected inaccessible, it will be restored by one of the other two copies immediately.
    \item \textbf{Documentation}: Our documentation will be updated regularly reflecting feedback from users.    
\end{itemize}

\newpage
\section{Dataset Organization}

After downloading and unzipping the zip files, please reorganize the files in the following tructure:
\begin{verbatim}
LAVN
   |--src
      |--makeData_virtual.py
      |--makeData_real.py
      ...
   |--Virtual
      |--Gibson
         |--traj_<SCENE_ID>
            |--worker_graph.json
            |--rgb_<FRAME_ID>.jpg
            |--depth_<FRAME_ID>.jpg
         |--traj_Ackermanville
            |--worker_graph.json
            |--rgb_00001.jpg
            |--rgb_00002.jpg
            ...
            |--depth_00001.jpg
            |--depth_00002.jpg
            ...
         ...
      |--Matterport
         |--traj_<SCENE_ID>
            |--worker_graph.json
            |--rgb_<FRAME_ID>.jpg
            |--depth_<FRAME_ID>.jpg
         |--traj_00000-kfPV7w3FaU5
            |--worker_graph.json
            |--rgb_00001.jpg
            |--rgb_00002.jpg
            ...
            |--depth_00001.jpg
            |--depth_00002.jpg
            ...
         ...
   |--Real
      |--Campus
         |--worker_graph.json
         |--traj_480p_<SCENE_ID>
            |--rgb_<FRAME_ID>.jpg
         |--traj_480p_scene00
            |--rgb_00001.jpg
\end{verbatim}
where the main landmark annotation scripts \textit{makeData\_virtual.py} and \textit{makeData\_real.py} are in folder (1) \textit{src}. (2) \textit{Virtual} and (3) \textit{Real} store trajectories collected in the simulation and real world, respectively. Each trajectory's data is collected in the following format:
\begin{verbatim}
         |--traj_<SCENE_ID>
            |--worker_graph.json
            |--rgb_<FRAME_ID>.jpg
            |--depth_<FRAME_ID>.jpg
\end{verbatim}
where \textit{$<SCENE\_ID>$} matches exactly the original one in \href{https://github.com/StanfordVL/GibsonEnv/blob/master/gibson/data/README.md}{Gibson} and \href{https://aihabitat.org/datasets/hm3d/}{Matterport} run by the photo-realistic simulator \href{https://github.com/facebookresearch/habitat-sim}{Habitat}. Images are saved in either \textit{.jpg} or \textit{.png} format. Note that \textit{rgb} images are the main visual representation while \textit{depth} is the auxiliary visual information captured only in the virtual environment. Real-world RGB images are downsampled to a \textit{$640 \times 480$} resolution noted by \textit{480p} in a trajectory folder name.

\textit{worker\_graph.json} stores the meta data in dictionary in Python saved in \textit{json} file with the following format:

\begin{verbatim}
{"node<NODE_ID>":
  {"img_path": "./human_click_dataset/traj_<SCENE_ID>/rgb_<FRAME_ID>.jpg",
   "depth_path": "./human_click_dataset/traj_<SCENE_ID>/depth_<FRAME_ID>.png",
   "location": [<LOC_X>, <LOC_Y>, <LOC_Z>],
   "orientation": <ORIENT>,
   "click_point": [<COOR_X>, <COOR_Y>],
   "reason": ""},
  ...
 "node0":
  {"img_path": "./human_click_dataset/traj_00101-n8AnEznQQpv/rgb_00002.jpg",
   "depth_path": "./human_click_dataset/traj_00101-n8AnEznQQpv/depth_00002.jpg",
   "location": [0.7419548034667969, -2.079209327697754, -0.5635206699371338],
   "orientation": 0.2617993967423121,
   "click_point": [270, 214],
   "reason": ""}
 ...
 "edges":...
 "goal_location": null,
 "start_location": [<LOC_X>, <LOC_Y>, <LOC_Z>],
 "landmarks": [[[<COOR_X>, <COOR_Y>], <FRAME_ID>], ...],
 "actions": ["ACTION_NAME", "turn_right", "move_forward", "turn_right", ...]
 "env_name": <SCENE_ID>
}
\end{verbatim}
where \textit{$[<LOC\_X>, <LOC\_Y>, <LOC\_Z>]$} is the 3-axis location vector, \textit{$<ORIENT>$} is the orientation only in simulation. \textit{$[<COOR\_X>, <COOR\_Y>]$} are the image coordinates of landmarks. \textit{ACTION\_NAME} stores the action of the robot take from the current frame to the next frame.

\section{{Dataset Usage}}
{This section highlights two detailed examples of efficient visual navigation training. The visual navigation task can be formulated as various types of problems, including but not limited to:}

{\textbf{1. Supervised Learning} by mapping visual observations (\textit{RGBD}) to waypoints (image coordinates). A developer can design a vision network whose input (\textit{$X$}) is \textit{RGBD} and output (\textit{$Y$}) is image coordinate, specified by \textit{$img\_path$}, \textit{$depth\_path$} and \textit{$click\_point$} \textit{$[<COOR\_X>, <COOR\_Y>]$} in the worker \textit{graph.json} file in the dataset. The loss function can be designed to minimize the discrepancy between the predicted image coordinate (\textit{$Y\_pred$}) and the ground truth (\textit{$Y$}), e.g. $loss = ||Y\_pred - Y||$. Then $Y\_pred$ can be simply translated to a robot’s moving action, such as $Y\_pred$ in the center or top region of an image means moving forward while $left/right$ regions represent turning left or right.}

{\textbf{2. Map Representation Learning} in the latent space of a neural network. One can train this latent space to represent two observations’ proximity by contrastive learning. The objective is to learn a function $h()$ that predicts the distance given two observations ($X_{1}$) and ($X_{2}$): $dist = h(X_{1}, X_{2})$. Note that $dist()$ can be a cosine or distance-based function, depending on the design choice. The positive samples can be nodes (a node includes information at a timestep such as \textit{RGBD} data and image coordinates) nearby while further nodes can be treated as negative samples. A landmark is a sparse and distinct object or scene in the dataset that facilitates a more structured and global connection between nodes, which further assists in navigation in more complex or longer trajectories.}

\newpage
\section{Extended Dataset Details}

\begin{figure}[h]
  \centering
  \includegraphics[width=0.96\textwidth]{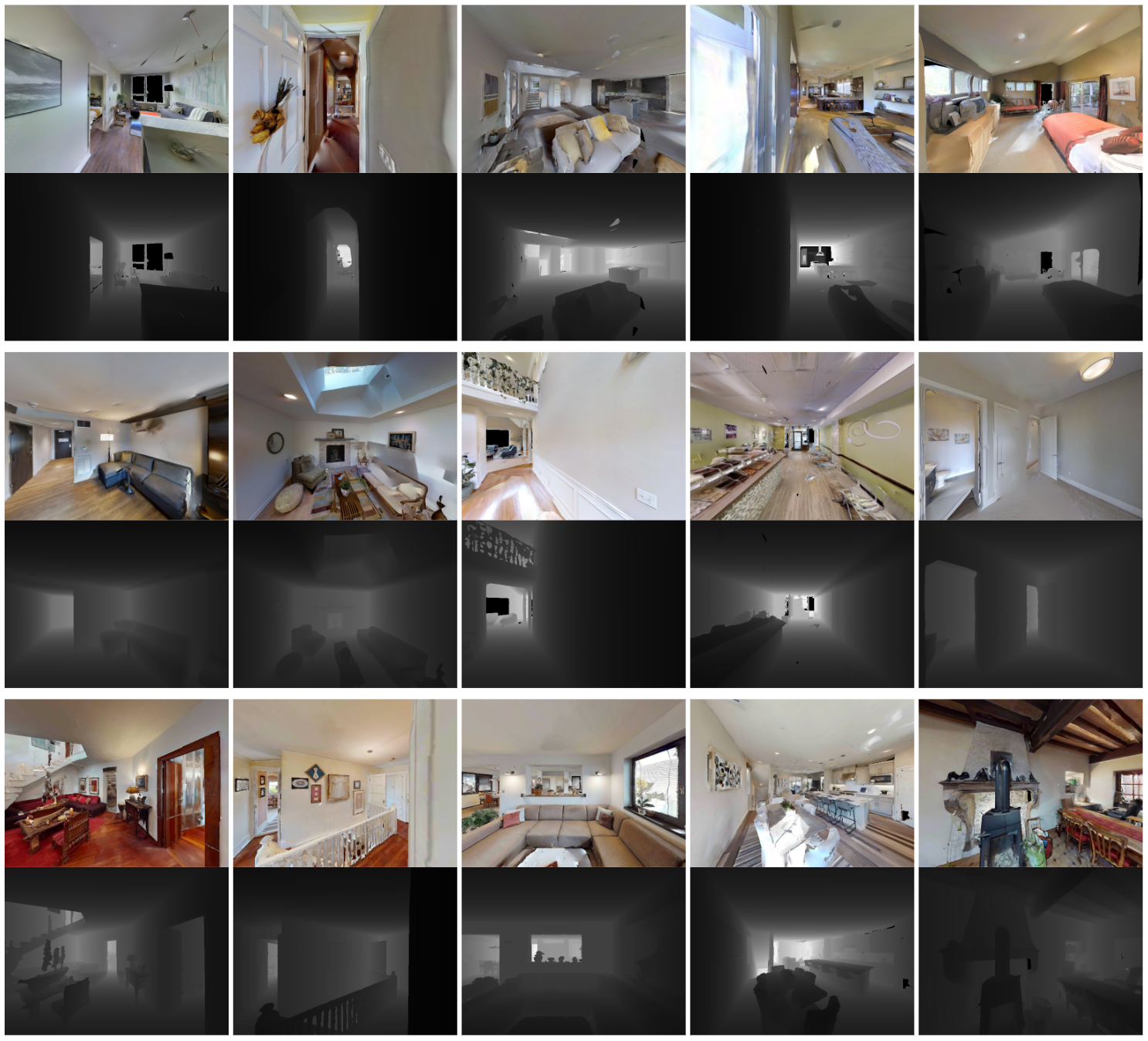}
  \caption{Samples of pairs of RGB and depth images in virtual environments in two rows.}
  \label{fig:dataset_depth}
\end{figure}

\begin{figure}[h]
  \centering
  \includegraphics[width=0.75\linewidth]{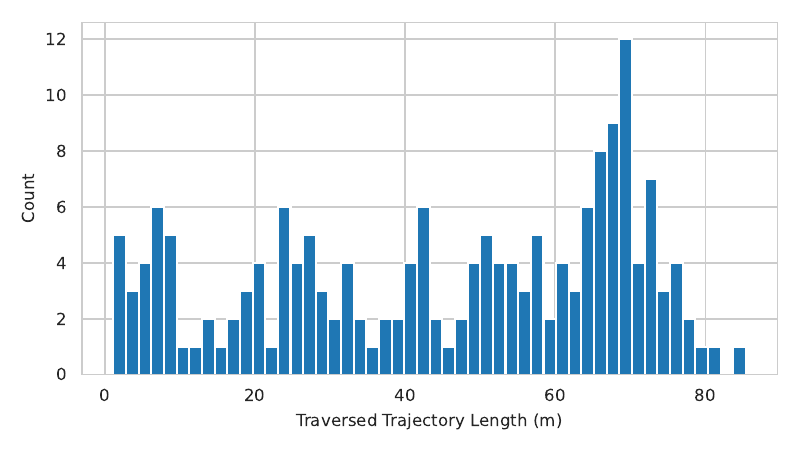}
  \caption{Histogram of traversed trajectory lengths in all virtual scenes in meters (m). LAVN consists of trajectories in diverse lengths.}
  \label{fig:traj_len}
\end{figure}

{We aim to randomly select rooms, evenly distributed by room IDs, to ensure diversity in room types and sizes, e.g. covering Room ID starting with letter A-Z in the Gibson dataset, 
digits 000-007 in the Matterport. This randomness results in diverse sets of rooms without bias in terms of room type (restroom, kitchen, hallway, office, garage, etc. in Fig. \ref{fig:dataset_depth}) and room size (shown in the diverse traversed trajectory length ranging from around 5 to 85 meters in Fig. \ref{fig:traj_len}).}

We present the full list of scene IDs in Gibson \cite{xiazamirhe2018gibsonenv} in Table \ref{tab:gibson_scene_ids}.

\begin{table}[h]
  \begin{center}
  \small
\begin{tabular}{p{9em}p{9em}p{9em}p{9em}}
\toprule
Ackermanville & Adairsville & Adrian & Airport \\
Albertville & Aldrich & Alfred & Allensville \\
Almena & Almota & Aloha & Alstown \\
American & Anaheim & Ancor & Andover \\
Angiola & Annawan & Annona & Anthoston \\
Apache & Applewold & Arbutus & Archer \\
Arkansaw & Arona & Artois & Ashport \\
Assinippi & Athens & Auburn & Aulander \\
Avonia & Azusa & Ballantine & Ballou \\
Baneberry & Barboursville & Barranquitas \\
Bautista & Beach & Beechwood & Bellwood \\
Benevolence & Bethlehem & Bettendorf & Biltmore \\
Blackstone & Bohemia & Bolton & Bonesteel \\
Bonnie & Booth & Bountiful & Bowlus \\
Bowmore & Branford & Braxton & Bremerton \\
Brevort & Brewton & Broseley & Brown \\
Browntown & Burien & Bushong & Byers \\
Cabin & Callicoon & Calmar & Cantwell \\
Capistrano & Carpendale & Carpio & Castor \\
Castroville & Channel & Checotah & Chesterbrook \\
Chilhowie & Chiloquin & Chireno & Churchton \\
Circleville & Cisne & Clive & Cobalt \\
Codell & Coffeen & Collierville & Convoy \\
Cooperstown & Copemish & Cottonport & Country \\
Crandon & Crookston & Culbertson & Cullison \\
Cutlerville & Dansville & Darden & Darnestown \\
Deatsville & Dedham & Deemston & Delton \\
Denmark & Dryville & Duarte & Duluth \\
Eagan & Eagerville & Eastville & Edgemere \\
Ellaville & Elmira & Elton & Emmaus \\
Eudora & Euharlee & Everton & Ewell \\
Fedora & Galatia & Hacienda & Idanha \\
Jenners & Kangley & Lajas & MacArthur \\
Macedon & Neibert & Nemacolin & Ogilvie \\
Okabena & Pablo & Pasatiempo & Quantico \\
Rancocas & Sagerton & Sanctuary & Tallmadge \\
Tansboro & Umpqua & Uncertain & Vails \\
Victorville & Voorhees & Waimea & Waipahu \\
Waldenburg & Yadkinville & Yankeetown & Yscloskey \\
\bottomrule
\end{tabular}
\end{center}
\caption{Full list of scene IDs in Gibson used in LAVN.}
\label{tab:gibson_scene_ids}
\end{table}

\newpage
The full list of scenes in Matterport \cite{Matterport3D} is listed in Table \ref{tab:matterport_scene_ids}:

\begin{table}[h]
  \begin{center}
\begin{tabular}{p{11em}p{11em}p{11em}p{11em}}
\toprule
00000-kfPV7w3FaU5 & 00001-UVdNNRcVyV1 & 00002-FxCkHAfgh7A & 00003-NtVbfPCkBFy \\
00004-VqCaAuuoeWk & 00005-yPKGKBCyYx8 & 00007-UQuchpekHRJ & 00008-VYnUX657cVo \\
00010-DBjEcHFg4oq & 00011-1W61QJVDBqe & 00012-kDgLKdMd5X8 & 00013-sfbj7jspYWj \\
00014-nYYcLpSzihC & 00015-LPwS1aEGXBb & 00017-oEPjPNSPmzL & 00018-as8Y8AYx6yW \\
00019-AfKhsVmG8L4 & 00020-XYyR54sxe6b & 00021-yQESfVcg18k & 00022-gmuS7Wgsbrx \\
00023-zepmXAdrpjR & 00031-Wo6kuutE9i7 & 00032-jTTGECZYKRA & 00033-oPj9qMxrDEa \\
00034-6imZUJGRUq4 & 00035-3XYAD64HpDr & 00036-41FNXLAZZgC & 00037-oKFJo8jpzRW \\
00038-aJg466zMSNt & 00039-ANmWrL7Kz7h & 00040-ZB8o8rMmPdB & 00102-r77mpaAYUEc \\
00103-gUqgeUmUagL & 00104-KJxdMPgweZG & 00105-xWvSkKiWQpC & 00106-ZVScmfktNQ1 \\
00107-Y6WjWkVEUks & 00109-GTV2Y73Sn5t & 00111-AMEM2eWycTq & 00112-r38SGhq8aJr \\
00113-3goH1WRaCYC & 00114-Coer9RdivP7 & 00115-NBWrHFXBF5p & 00116-xp4FyfQ6Wr5 \\
00117-2NwLiyeKcrK & 00118-F5j7ZLfMm1n & 00120-eAUmfFLZDR3 & 00121-D2PqRE5ZvyQ \\
00122-QDtpZSqaeyW & 00123-C3ifY177Ldq & 00201-k7vRbGpz44m & 00202-yVbpFay8gTU \\
00203-VoVGtfYrpuQ & 00204-gxttMtT5ZGK & 00205-NEVASPhcrxR & 00206-uhkqDVMtEnn \\
00207-FRQ75PjD278 & 00209-C5RbHBQ76DE & 00210-j2EJhFEQGCL & 00211-hmRxh2mmzNC \\
00212-bAdy4hKf1a1 & 00213-mkvHBa3mEEk & 00214-WeyCwVzL53K & 00215-zWydhyFhvcj \\
00216-6EMViBCA2N7 & 00217-qz3829g1Lzf & 00218-fQHGxvurx9L & 00220-KAzjXJvZtR3 \\
00221-zJEEFaNaRbB & 00222-g8Xrdbe9fir & 00301-JiHGQpwKUvd & 00302-JFgrz9MNz4b \\
00303-ghWQ5kHV97i & 00304-X6Pct1msZv5 & 00305-W3J8ruZTQic & 00306-Y4L8fjz2yH7 \\
00307-vDfkYo5VqEQ & 00309-VKmpsujnc5t & 00310-WnvnMQh4eEa & 00311-mHXUEKEV6gR \\
00312-UrFKpVJpvHi & 00313-PE6kVEtrxtj & 00314-NwG7cpZnRZb & 00315-We1N7vBtyGm \\
00316-LqsTKpxKVP2 & 00317-P8XJUpcAUkf & 00318-6qJyEsZNuey & 00319-sjH1uaR68XQ \\
00320-nicaPonCxvC & 00321-JWWJBQWHv64 & 00401-H8rQCnvBgo6 & 00402-zR6kPe1PsyS \\
00404-QN2dRqwd84J & 00406-n2Tt2eJdqnT & 00407-NPHxDe6VeCc & 00408-RamZzGBBPbT \\
00409-rxGLNxH6eoJ & 00410-v7DzfFFEpsD & 00411-o4tckGBtaxz & 00412-mDPCxA7W1WN \\
00413-YM4nG4pSAEJ & 00414-77mMEyxhs44 & 00415-rBmEe6ab5VP & 00416-zCMdfYaW9iF \\
00417-nGhNxKrgBPb & 00419-fbUcgfPMBDr & 00420-R6Byftz8wRN & 00421-gamLwhSzHci \\
00422-8wJuSPJ9FXG & 00423-bEdki9cbHDG & 00501-N7YVmJQ8sAu & 00502-nMeXfQU4PMS \\
00504-frThKkhTwFT & 00505-ZwnLFNzxASM & 00506-QVAA6zecMHu & 00507-RfNGMBdVbAZ \\
00508-4vwGX7U38Ux & 00509-gDDiZeyaVc2 & 00510-JSgMy8tTACD & 00511-8uSpPmctPXC \\
00601-PjnDyQJJ3eM & 00604-W4r5JssudHR & 00605-T22dejNjHK7 & 00606-W16Bm4ysK8v \\
00607-JXdzHne1mRo & 00608-j2Nms3h9XJv & 00609-x1pTUWx9DPr & 00611-PXAfUkZGMdU \\
00612-GsQBY83r3hb & 00613-s19Uyn7AWwv & 00614-ki6Cu76pWzF & 00615-PUNuHY5M7MS \\
00616-zhzot8MvSjF & 00617-AENiMBDjVFb & 00618-T4G9hTR5WSv & 00619-R9fYpvCUkV7 \\
00620-AUkcTmUs8mw & 00622-bxwHR9ipFG8 & 00701-tpxKD3awofe & 00702-wCqnXzoru3X \\
00703-xvDx98avcwd & 00704-qnKYFQsjnHf & 00705-2XVvKEDd54w & 00706-YHmAkqgwe2p \\
00707-XVSZJAtHKdi & 00708-eUJx9a4u63E & 00709-8LLjiNrWzJ9 & 00710-DGXRxHddGAW \\
\bottomrule
\end{tabular}
\end{center}
\caption{Full list of scene IDs in Matterport used in LAVN.}
\label{tab:matterport_scene_ids}
\end{table}

\newpage

We present RGB observations and the corresponding depth image samples in Fig. \ref{fig:dataset_depth}.

\newpage
Statistics of trajectory lengths that a robot has traversed over all virtual environments recorded in meters (m) is shown in Fig. \ref{fig:traj_len}. The wide spread of trajectory lengths (ranging from 1.16m to 85.38m) provides a significant temporal diversity that enriches a robot's learning recipe with both short- and long-term dependencies.

\newpage
Trajectories are recorded in ten diverse scenes in both indoor and outdoor environments shown in Fig. \ref{fig:dataset}.

\begin{figure}[h]
  \centering
  \includegraphics[width=0.85\textwidth]{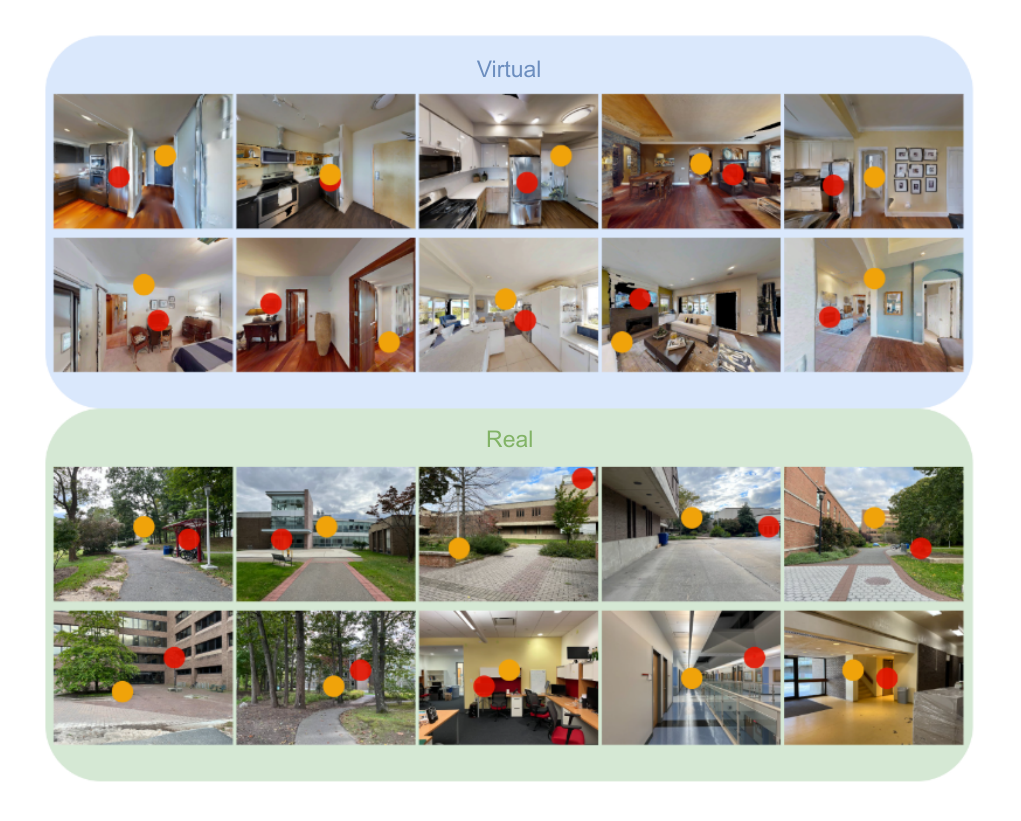}
  \caption{Samples of RGB images in virtual (1st and 2nd rows in blue) and real-world (3rd and 4th rows in green) environments with landmarks and waypoints annotated by red and orange dots, respectively. Trajectories are recorded in comprehensive environments both indoors and outdoors.}
  \label{fig:dataset}
\end{figure}

\newpage
\section{Datasheets for Datasets}
We present the answers to the questions of datasheets for datasets in this sub-sections.
\subsection{Motivation}
\begin{itemize}
    \item \textbf{For what purpose was the dataset created?}
    \textbf{Answer:} \ds~ is created to assist efficient representation learning in embodied visual navigation.  
    \item \textbf{Who created the dataset (e.g., which team, research group) and on behalf of which entity (e.g., company, institution, organization)?} \textbf{Answer:} The authors in this paper.
    \item \textbf{Who funded the creation of the dataset? If there is an associated grant, please provide the name of the grantor and the grant name and number.} \textbf{Answer:} We will release the funding source upon acceptance.
\end{itemize}

\subsection{Composition}
\begin{itemize}
    \item \textbf{What do the instances that comprise the dataset represent (e.g., documents, photos, people, countries)?} \textbf{Answer:} LAVN consists of RGB and depth images, landmark and waypoint annotations saved in image coordinate formats in a json files. The collected environments include virtual in Habitat AI \cite{habitat19iccv} and real world scenes without any human involvement. The details are presented in the main paper.
    \item \textbf{How many instances are there in total (of each type, if appropriate)?} \textbf{Answer:} There are 310 trajectories, 103,998 frames (RGB and depth images), 14,281 landmarks in 300 virtual and 10 real environments in this dataset.
    \item \textbf{Does the dataset contain all possible instances or is it a sample (not necessarily random) of instances from a larger set?} \textbf{Answer:} It contains all possible instances.
    \item \textbf{What data does each instance consist of?} \textbf{Answer:} Each instance consists of an RGB and depth image, the corresponding waypoint in image coordinate annotated by human that indicates the direction to move, the corresponding action inferred by the waypoint as well as the landmark in image coordinate if recognized by a human.
    \item \textbf{Is there a label or target associated with each instance?} \textbf{Answer:} Yes, the label includes the waypoint, landmark if exists.
    \item \textbf{Is any information missing from individual instances? If so, please provide a description, explaining why this information is missing (e.g., because it was unavailable). This does not include intentionally removed information, but might include, e.g., redacted text.} \textbf{Answer:} N/A.
    \item \textbf{Are relationships between individual instances made explicit (e.g., users’ movie ratings, social network links)?} \textbf{Answer:} N/A.
    \item \textbf{Are there recommended data splits (e.g., training, development/validation, testing)?} \textbf{Answer:} No. Users may choose any train/val split strategies.
    \item \textbf{Are there any errors, sources of noise, or redundancies in the dataset?} \textbf{Answer:} We checked the correctness of our dataset. If such errors exist, we will further correct our dataset.
    \item \textbf{Is the dataset self-contained, or does it link to or otherwise rely on external resources (e.g., websites, tweets, other datasets)?} \textbf{Answer:} It is self-contained. Users can use it directly without the use of the virtual environment such as Habitat AI \cite{habitat19iccv}.
    \item \textbf{Does the dataset contain data that might be considered confidential (e.g., data that is protected by legal privilege or by doctor–patient confidentiality, data that includes the content of individuals’ non-public communications)?} \textbf{Answer:} N/A.
    \item \textbf{Does the dataset contain data that, if viewed directly, might be offensive, insulting, threatening, or might otherwise cause anxiety?} \textbf{Answer:} No.
    \item \textbf{Does the dataset relate to people?} \textbf{Answer:} No.
    \item \textbf{Does the dataset identify any subpopulations (e.g., by age, gender)?} \textbf{Answer:} No.
    \item \textbf{Is it possible to identify individuals (i.e., one or more natural persons), either directly or indirectly (i.e., in combination with other data) from the dataset?} \textbf{Answer:} N/A.
    \item \textbf{Does the dataset contain data that might be considered sensitive in any way (e.g., data that reveals race or ethnic origins, sexual orientations, religious beliefs, political opinions or union memberships, or locations; financial or health data; biometric or genetic data; forms of government identification, such as social security numbers; criminal history)?} \textbf{Answer:} No.
\end{itemize}

\subsection{Collection Process}
\begin{itemize}
    \item How was the data associated with each instance acquired?
    \textbf{Answer:} For virtual environments, we run the simulator Habitat AI on a desktop. In each step, it displays the current frame and a human annotates the waypoint by left click or landmark (if exists) by right click on the frame. The robot will move according to the action occurred by waypoint. Then the corresponding data (RGB, depth, waypoint and landmark) will be saved on disk. For real-world scenes, a human use a cellphone's camera acting as a robot to traverse in the environment. Real-world data collection process follows the same procedure in the virtual environment, except that waypoints and landmarks are annotated during post-processing of videos.
    \item \textbf{What mechanisms or procedures were used to collect the data (e.g., hardware apparatuses or sensors, manual human curation, software programs, software APIs)?} \textbf{Answer:} We use Habitat AI \cite{habitat19iccv} run on a desktop in Ubuntu 20.04 LST equipped with two NVIDIA RTX 1080 GPUs. An iPhone 13 mini \cite{iphone13mini} is used to collect real-world data.
    \item \textbf{If the dataset is a sample from a larger set, what was the sampling strategy (e.g., deterministic, probabilistic with specific sampling probabilities)?} \textbf{Answer:} For virtual data, we collect images in Gibson \cite{xiazamirhe2018gibsonenv} and Matterport \cite{Matterport3D} datasets in a deterministic way.
    \item \textbf{Who was involved in the data collection process (e.g., students, crowdworkers, contractors) and how were they compensated (e.g., how much were crowdworkers paid)?} \textbf{Answer:} The authors of this paper volunteered to help with the data collection process.
    \item \textbf{Over what timeframe was the data collected?} \textbf{Answer:} Data was collected in September 2023.
    \item \textbf{Were any ethical review processes conducted (e.g., by an institutional review board)?} \textbf{Answer:} N/A. Not required as no human involved.
    \item \textbf{Does the dataset relate to people?} \textbf{Answer:} No.
\end{itemize}

\subsection{Preprocessing/cleaning/labeling}
\begin{itemize}
    \item \textbf{Was any preprocessing/cleaning/labeling of the data done (e.g., discretization or bucketing,
tokenization, part-of-speech tagging, SIFT feature extraction, removal of instances, processing of missing values)?} \textbf{Answer:} We removed trajectories that contain incorrect labels (waypoints or landmarks).
    \item \textbf{Was the “raw” data saved in addition to the preprocessed/cleaned/labeled data (e.g., to support unanticipated future uses)?} \textbf{Answer:} Yes.
    \item \textbf{Is the software that was used to preprocess/clean/label the data available?} \textbf{Answer:} We use Google Sheets and Python scripts written by ourselves.
\end{itemize}

\subsection{Uses}
\begin{itemize}
    \item \textbf{Has the dataset been used for any tasks already?} \textbf{Answer:} Yes, for improving embodied visual navigation representation learning.
    \item \textbf{Is there a repository that links to any or all papers or systems that use the dataset?} \textbf{Answer:} We will release the papers and systems that use the dataset upon acceptance.
    \item \textbf{What (other) tasks could the dataset be used for?} \textbf{Answer:} N/A.
    \item \textbf{Is there anything about the composition of the dataset or the way it was collected and preprocessed/cleaned/labeled that might impact future uses?} \textbf{Answer:} Combing RGB and depth images with landmarks may help improve visual navigation in the future.
    \item \textbf{Are there tasks for which the dataset should not be used?} \textbf{Answer:} No.
\end{itemize}

\subsection{Distribution}
\begin{itemize}
    \item \textbf{Will the dataset be distributed to third parties outside of the entity (e.g., company, institution, organization) on behalf of which the dataset was created?} \textbf{Answer:} No.
    \item \textbf{How will the dataset will be distributed (e.g., tarball on website, API, GitHub)?} \textbf{Answer:} Yes. We release our dataset at \href{https://huggingface.co/datasets/visnavdataset/lavn}{https://huggingface.co/datasets/visnavdataset/lavn}.
    \item \textbf{When will the dataset be distributed?} \textbf{Answer:} Already distributed.
    \item \textbf{Will the dataset be distributed under a copyright or other intellectual property (IP) license, and/or under applicable terms of use (ToU)?} \textbf{Answer:} The dataset is distributed under MIT License.
    \item \textbf{Have any third parties imposed IP-based or other restrictions on the data associated with the instances?} \textbf{Answer:} No.
    \item \textbf{Do any export controls or other regulatory restrictions apply to the dataset or to individual instances?} \textbf{Answer:} No.
\end{itemize}

\subsection{Maintenance}
\begin{itemize}
    \item \textbf{Who will be supporting/hosting/maintaining the dataset?} \textbf{Answer:} The authors of this paper.
    \item \textbf{How can the owner/curator/manager of the dataset be contacted (e.g., email address)?} \textbf{Answer:} The authors of this paper. The email address will be released upon acceptance.
    \item \textbf{Is there an erratum?} \textbf{Answer:} No.
    \item \textbf{Will the dataset be updated (e.g., to correct labeling errors, add new instances, delete instances)?} \textbf{Answer:} If such update is needed, we will correct the dataset and release a new version.
    \item \textbf{If the dataset relates to people, are there applicable limits on the retention of the data associated with the instances (e.g., were the individuals in question told that their data would be retained for a fixed period of time and then deleted)?} \textbf{Answer:} N/A. This dataset is not related to people.
    \item \textbf{Will older versions of the dataset continue to be supported/hosted/maintained?} \textbf{Answer:} We plan to maintain the latest version.
    \item \textbf{If others want to extend/augment/build on/contribute to the dataset, is there a mechanism for them to do so?} \textbf{Answer:} Individual contributors can contact the authors of this paper.
\end{itemize}

\end{document}